# LSTM vs. GRU vs. Bidirectional RNN for script generation

Sanidhya Mangal
Computer Science and Engineering
*Medi-Caps University*
Indore, India
mangalsanidhya19@gmail.com

Poorva Joshi
Computer Science and Engineering
*Medi-Caps University*
Indore, India
purvaj27@gmail.com

Rahul Modak
Computer Science and Engineering
*Medi-Caps University*
Indore, India
rahulsvmodak@gmail.com

***Abstract*— Scripts are an important part of any TV series. They narrate movements, actions and expressions of characters. In this paper, a case study is presented on how different sequence to sequence deep learning models perform in the task of generating new conversations between characters as well as new scenarios on the basis of a script (previous conversations). A comprehensive comparison between these models, namely, LSTM, GRU and Bidirectional RNN is presented. All the models are designed to learn the sequence of recurring characters from the input sequence. Each input sequence will contain, say 'n' characters, and the corresponding targets will contain the same number of characters, except, they will be shifted one character to the right. In this manner, input and output sequences are generated and used to train the models. A closer analysis of explored models' performance and efficiency is delineated with the help of graph plots and generated texts by taking some input string. These graphs describe both, intraneural performance and interneural model performance for each model.**

## I. Introduction

A script contains dialogues of the characters and also a description of scenes that appear. It is a formal way of structuring dialogues that occur in a play, TV show, movie etc. Script narrates movements, actions and expressions of characters.

In this paper, we have presented a case study about how different deep learning models perform in the task of generating new conversations between characters as well as new scenarios on the basis of a script (previous conversations). As this task is similar to text generation, different sequence to sequence text generation models are used in this paper. The models for text generation are trained on a sequence of characters in our text to predict the next character in that sequence. Each input sequence will contain, say n characters, and the corresponding targets will contain the same number of characters, except, they will be shifted one character to the right. In this manner, we will generate our input and output sequences and use them to train our models. In this paper, three types of recurrent neural networks are used, namely, Bidirectional Recurrent Neural Networks (BidirectionalRNN) [1], Long Short-Term Memory (LSTM) [2] and Gated Recurrent Units [3].

This paper gives a comprehensive comparison of the performances of these widely used text generation models used for this particular application of text generation.

Here, we used the script of a famous TV series which contains all the dialogues' and scenes' description of all the episodes crunched into a single pickle file [4]. When we consider a TV show which is long-running, it gives sufficient amount of data to train our neural models and now, this trained neural models can be used to get some idea about what the next episode of that TV show could be. This will assist the writers to generate a new episode using these predictions.

Additionally, this paper covers software implementation of proposed models and related work pursued by different researchers in this domain. Their work is cited and analyzed to produce stellar results for these models. In the latter sections of this paper, a closer analysis of models' performance and efficiency is discussed in detail along with outliers and proper plots of curves generated by training the models, along with the text generated after training these models.

## II. Related Work

Dialogue generation system is an antedate, many predecessors have explored this domain. Based on their work dialogue generation system can be built in several ways.

One of the ways is to treat dialogue generation as a source to target transduction problem and learns mapping rules between input messages and responses from a massive amount of training data as suggested by (Alan) [5].

Augmenting to above method a response generation problem is framed as statistical machine problem (SMT) Ritter et al (2011) [6]. Vinyals and Le, 2015, [7] incorporated [5] method to build end-to-end conversational systems which generate a response from message vector by applying encoder to map a message to a distributed vector resenting its semantics. Also, Li et al(2016a) [8] worked on the reduction of the proportion of generic responses produced by SEQ2SEQ (Sutskever et al) [9] systems. An end-to-end dialogue system using generative hierarchical neural network models is presented by Serban et al [10].

The other way is to build a task-oriented dialogue system to solve domain-specific tasks. Esther Levin [11] presented a stochastic model based on Markov Decision Process [12] to show that the problem of dialogue strategy design can be stated as in optimization problem. Partially Observable Markov Decision Processes (POMDP) by [13] presents a way to model uncertainty in dialogues using policy optimization based on grid-based Q-learning [14] with a summary of belief space. Su et al [15] proposed a model of continuously learning Neural Dialogue management.

Also, reinforcement learning can be used for the same purpose. Alan Ritter's paper [5] applies deep reinforcement learning (DL) to stimulate dialogues between two virtual agents, using the policy gradient method further rewarding chatbot dialogue.

## III. IMPLEMENTATION

This section embellishes the data which went under preprocessing and all the models that went under the process of training. A batch of fully trained models was used to generate text. Though the model training and experiment was conducted on Google Colab, running over Google Cloud Platform (GCP), the prediction task or text generation task was conducted on the local machine. Complete execution of model training and text prediction was done with the help of deep learning technology and code implementation on Keras (using Tensorflow as backend).

### *A. Deep Neural Network Design*

The model is applied to the multidimensional text vectors. Three different neural models (LSTM, GRU and Bidirectional RNN) were trained to imbibe the probability of occurrence of the next character in the sequence based on the current character. Neural Models are designed in such a way that they can retain the previous text up to 100-character sequences at a current step.

Neural models depend on selective inputs. Not every character sequence undergoes training process. Some selected and specified sequences are only used to train these models, which can effectively tune the model resulting in a maximum information gain. In all neural models, there is an Embedding [16] layer, which is the first hidden layer for all the models. This layer is useful in learning the dense mapping between the text data. Input in the layer of each neural model, whether it be LSTM, GRU or Bidirectional RNN, is derived from the output of Embedding layer to learn the mapping between character sequences and projection of these sequences. Next, to these layers, Dropout [17] layer is used to prevent the overfitting of data and provide more generalizations in the models by randomly switching on and off the neural units in these neural layers. All the neural layers (LSTM, GRU and Bidirectional RNN) with Dropout layer succeeding all these layers, in a combination of the uni-layered, bi-layered and quad-layered show the diversity of models. Once the model has learned the sequence mapping between all these character inputs, it is passed to a Dense [18] layer to combine all the neurons from the neuron layers, subordinating the gaps between these neurons. Dense layer ensures that all the neurons are fully connected with each other in a unison. In order to check how well or poorly a model is behaving after each optimization or training iteration, a loss value is used and, in this case, Sparse Categorical Cross Entropy [18] loss is used to ensure that performance of the model is maintained throughout the training process. The loss function is defined as:

$$\text{Loss} = -\frac{1}{N}\sum_{s\epsilon S}\sum_{c\epsilon C} 1_{s\epsilon c} logp(s\epsilon C)$$

Where S – samples, C – classes, s ϵ C – samples belongs to class c.

After training them, the neural models are ready to generate a new sequence of characters. To ensure better prediction and diverse output of sequences, an annotated dataset is used. The goal was to expose the model with a diverse dataset which would lead to a better tuning of the model. The text file format was used to extract dataset. The text files were annotated which played an important role in determining the speaker and the correct sequence of the dialogues.

The model was compiled using Moon et al. [19] as a suggested guide for dropouts. Dropout of 0.4 was applied to each of the neural layers, i.e., LSTM, GRU and Bidirectional LSTM. The optimizer selected was RMSprop [20], with the learning rate of 1e-3 for model parameter optimization.

### *B. Software Design*

Data is one of the most key components in training and validating any neural network. The data vector in this work is a sequence of characters which are converted into a sequence of number. A vocabulary dictionary named char2idx is used to map the unique characters into a unique set of numbers on which, a model can be trained. Similarly, in order to convert these numbers into corresponding characters, an idx2char dictionary is used for mapping all these unique numbers into a unique set of characters. Data vector, in this case, is referred to as “Sequence Matrix”, which is formed after taking a repeated sequence of characters converted into a number sequence. Before this Sequence Matrix can be used, a dataset undergoes a lot of preprocessing. Firstly, all the dataset from a text file is collected and serialized with the help of pickle [21] module. After this, the vocabularies of the dataset, char2idx and idx2char dictionaries are formed and with the help of these dictionaries and vocabularies, a sequence of characters is converted into the numerical sequence. Once a numerical sequence is generated with the help of TensorFlow dataset module, this array is converted into the Sequence Matrix and the dataset is split into two parts, the training features and the training labels, further shuffling and batching the dataset to add some variance and to improve model performance.

All the models as shown in the fig. 2 can be broken down into 5 function modules: 1) Embedding layer composed of 256 neurons (units) used to map all the unique vocabulary points into these 256 units. 2) A neural layer from the pool of LSTM, GRU and Bidirectional RNN with a combination of either single-layer, bi-layers, or quad-layers, a detailed description about the number of neurons in a particular layer is elucidated in Table 1. 3) Dropout layer, to generalize the learning process in the neural models to learn the sequences efficiently and prevent overfitting in these models. 4). Dense layer, this layer plays an important role in connecting all the neurons from neural layers in unison and produce the coveted output from these layers as per the need of the user. In this case, the output is similar to the input in the embedding layer, a vector of the dimension of vocabulary set. 5) Root Mean Square Propagation (RMSprop), this is similar to gradient descent [22] optimizer algorithm to optimize the steps during our model training so that the losses are converged at a faster rate. A short algorithm describing the functionality of RMSprop is presented below:

$$V_t = \rho V_{t-1} + (1 - \rho) * g_t^2$$

$$\Delta w_t = -\frac{\alpha}{\sqrt[2]{V_t + \epsilon}} * g_t$$

$$w_t = w_t + \Delta w_t$$

Where,

$$\alpha: Initial\ learning\ rate$$
$$V_t: Exponential\ average\ of\ squares\ of\ gradients$$
$$g_t: Gradient\ at\ time\ t\ along\ w^j$$

TABLE I. CONFIGURATION OF NEURONS IN EACH MODEL AND LAYERS

| **Model** | **Uni-layer** | **Bi-layers** | **Quad-layers** |
|---|---|---|---|
| LSTM | 1024 | 512,256 | 512, 256, 128, 64 |
| GRU | 1024 | 512, 256 | 512, 256, 128, 64 |
| Bidirectional RNN | 1024 | 512, 256 | 512, 256, 128, 64 |

Fig. 1. A table describing all the configuration on which the experiment was conducted. Entries in each cell of this table indicate the number of neurons used in each architecture of neural layers.

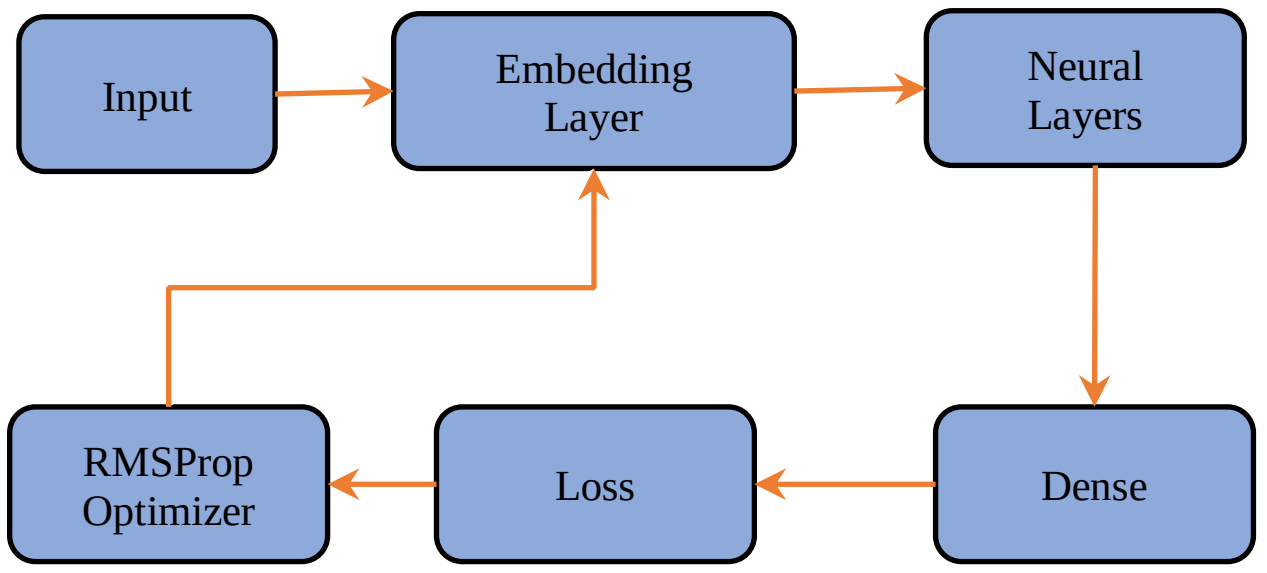

Fig. 2. The image describes architectural design used in this experiment, the model is represented in a form of end to end flow chart, beginning from the input layer and culminating at optimizer and loss step. In this image, a common block (Neural Layer) is used to represent all the configuration mentioned in Table 1.

The sparse categorial loss function is used to compute the losses to measure the dissimilarity between the distribution of actual labelled class and predicted the probability of the class membership. Categorial represents that the classes are not binary, i.e., there is a possibility of having more than two classes for a single input. Sparse refers to the classes that are not one-hot encoded. Instead, it is a single integer ranging from zero to the number of classes minus one. The sparse categorial loss function is proved to be beneficial when classes are mutually exclusive.

After the model is trained, it is stored in a serialized format to use it for the text generation. Weights from this serialized model are loaded into a slightly different configuration of the model, i.e., changing the dimensions of the model to process just a single character instead of a complete batch of the text sequence. Once the model is ready, a text sequence or a single character is used as an input to predict the upcoming character. This upcoming character is selected by dividing the probabilities of all these upcoming characters by a temperature. Lower temperature results in a more predictable text and higher temperature results in a more surprising text. For all the models in our experiment, the temperature is set to one. After the probabilities are normalized by temperature, the character with the highest probability is selected as an upcoming character. After the identification of character, it is augmented to the previous input and fed as input for the selection of the next character. This process is repeated for a specific number of iterations or word counts. Fig. 3 describes how does generation function works and prediction of new characters done.

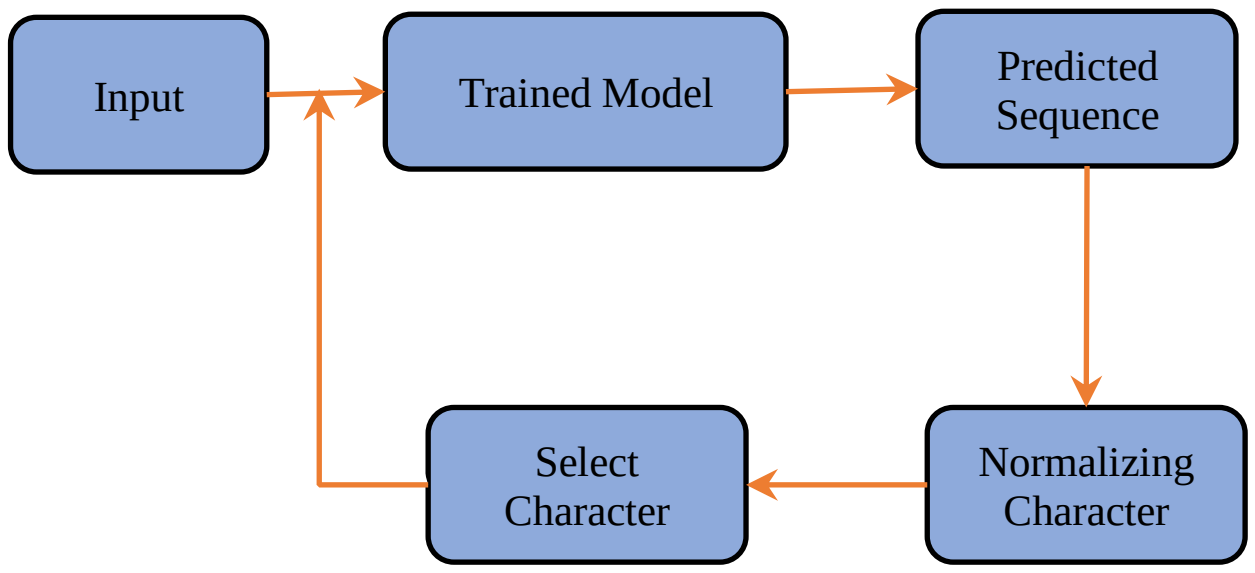

Fig. 3. This image describes the complete process that takes place when a text generation function is invoked to generate text when a model is trained.

Link to the code of this work, trained models of LSTM, GRU and Bidirectional RNN for all the configuration [23]. The output from each model in a text file along with the training history for all the models are also present in a form of pickle file.

## IV. RESULTS

Fig. 4 shows a comparative study between all the variants of the neural layers to train and generate text sequence for the scripts, as well as the average time required to train each of these steps on Google Colab running over GCP. While training different models, it was observed that LSTM based neural networks took the least time to execute a training epoch, Bidirectional RNN took the most time and GRU took slightly greater time than LSTM. An important factor affecting the performance of these models was training data set as there was a limited amount of structured data available publicly for any single tv series. To expose these models to the highest variation of a text sequence, the annotated format of data was taken from [24] repo publicly available on GitHub.

TABLE II. RESULT ANALYSIS OF DIFFERENT MODELS

| Model | Log-Likelihood | Training Step Time (ms) |
|---|---|---|
| LSTM (single layer) | -0.064 | 28 |
| LSTM (bi-layer) | 0.170 | 24 |
| LSTM (quad-layer) | 0.394 | 17 |
| GRU (single layer) | 0.080 | 24 |
| GRU (bi-layer) | 0.216 | 17 |
| GRU (quad-layer) | 0.423 | 14 |
| Bidirectional RNN (single layer) | -2.928 | 108 |
| Bidirectional RNN (bi-layer) | -3.216 | 62 |
| Bidirectional RNN (quad-layer) | -2.535 | 49 |

Fig. 4. This table represents the overall performance for each architecture of the model. The table contains two columns the first column represents mean log-likelihood loss for the given model and second column represent the time taken by each batch iteration in milliseconds to execute on Google Colab.

More intuition concerning the performance of various neural layers can be gained by looking at the graphs obtained after training each of them recursively on the dataset.

Fig. 5 delineates the intraneural layer performance for all the variants of the neural models whether it be LSTM, GRU or Bidirectional RNN. Plots in this figure draw an analogy between single layer, bi-layer and quad-layer configuration for all the neural models. It can be deciphered from fig. 5(a) that for all the configurations of LSTM layers, the losses at the initial state were approximately the same, ranging from 2.4 to 3.2. A hyperbolic curve is observed when the loss after every single iteration is plotted, with all the configurations showing a massive drop in the loss till 9$^{th}$ iteration. From the 10$^{th}$ iteration, the loss tends to become non-aberrant to some extent. In both the cases, i.e., bi-layered LSTM and quad-layered LSTM, the losses are ebbing till the last iteration(75$^{th}$). But in case of single-layered LSTM architecture, the loss drops till 30$^{th}$ iteration and gradually starts to ascent. Despite the increase in the loss of single-layered architecture, the loss remains minimum in this case with just 88E-2. In bi-layered and quad-layered architecture, the losses culminate at 107E-2 and 126E-2 respectively. On taking a look at fig. 5(b), it can be deduced that the losses for the single-layered GRU model and bi-layered model are aggregated near 240E-2. But in case of the quad-layered model, the loss bumps up to nearly 300E-2. Unlike LSTM architecture, the chute in loss is observed till 10$^{th}$ iteration, but after that, the drop becomes more unwavering. In the case of GRUs as well, there are some anomalies detected for single-layered architecture. Loss tends to drop till 25$^{th}$ iteration touching the value of 96E-2, later showing an ascent in the value of loss ceasing at 116E-2. Fig. 5(c) describes the losses over epochs for all configurations of Bidirectional RNN architecture. It can be construed that during 1$^{st}$ iteration, losses for single-layered and bi-layered architecture are agglomerated around 80E-2 but quad-layer shows an aberration going up as high as 259E-2. Unlike other neural architecture, Bidirectional RNN model converges right after single iteration, with the values 08E-2, 06E-2 and 21E-2 for single layer, bi-layer and quad-layer respectively. From 2$^{nd}$ iteration, losses begin to drop gradually and steadily. It can be observed from fig 5(c) that for all the configurations, the losses do not show much deviation and halt at 32E-3, 26E-3 and 37E-3 for single, bi and quad architectures respectively.

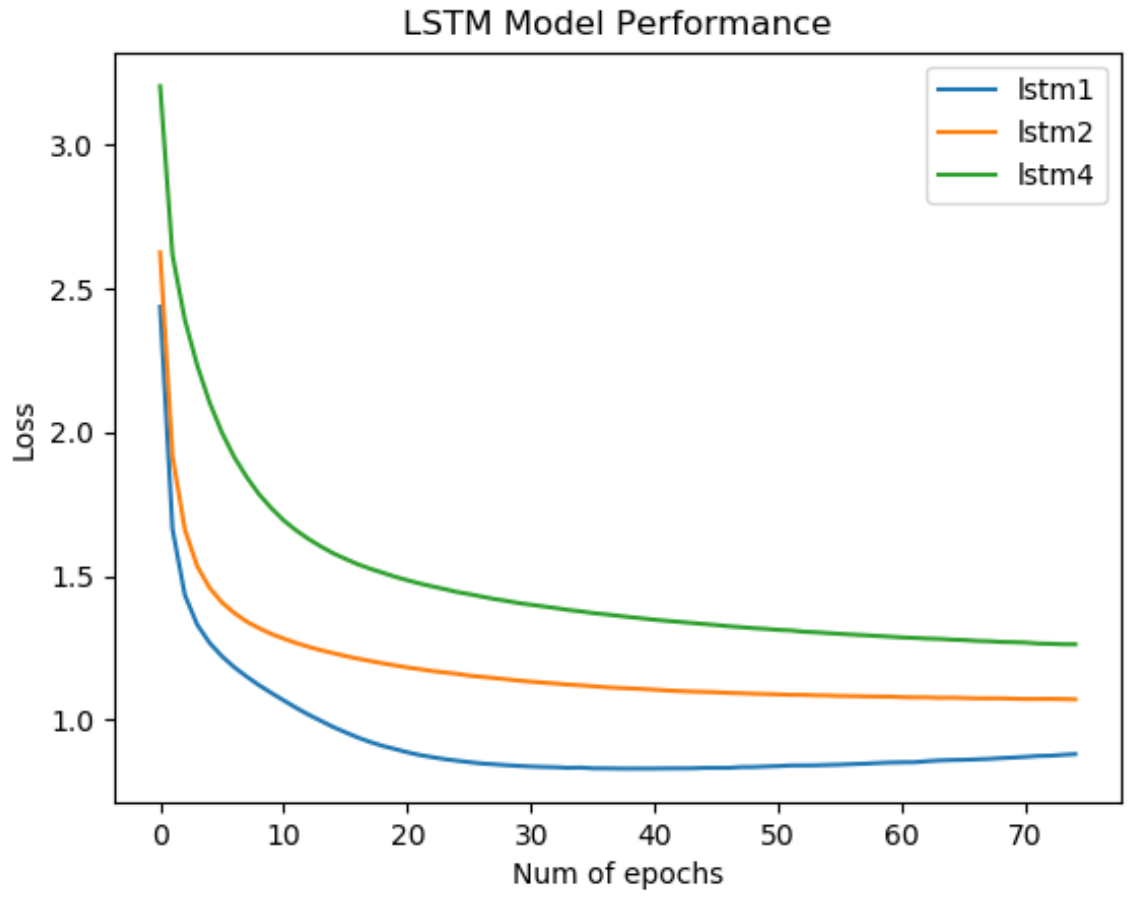

(a)

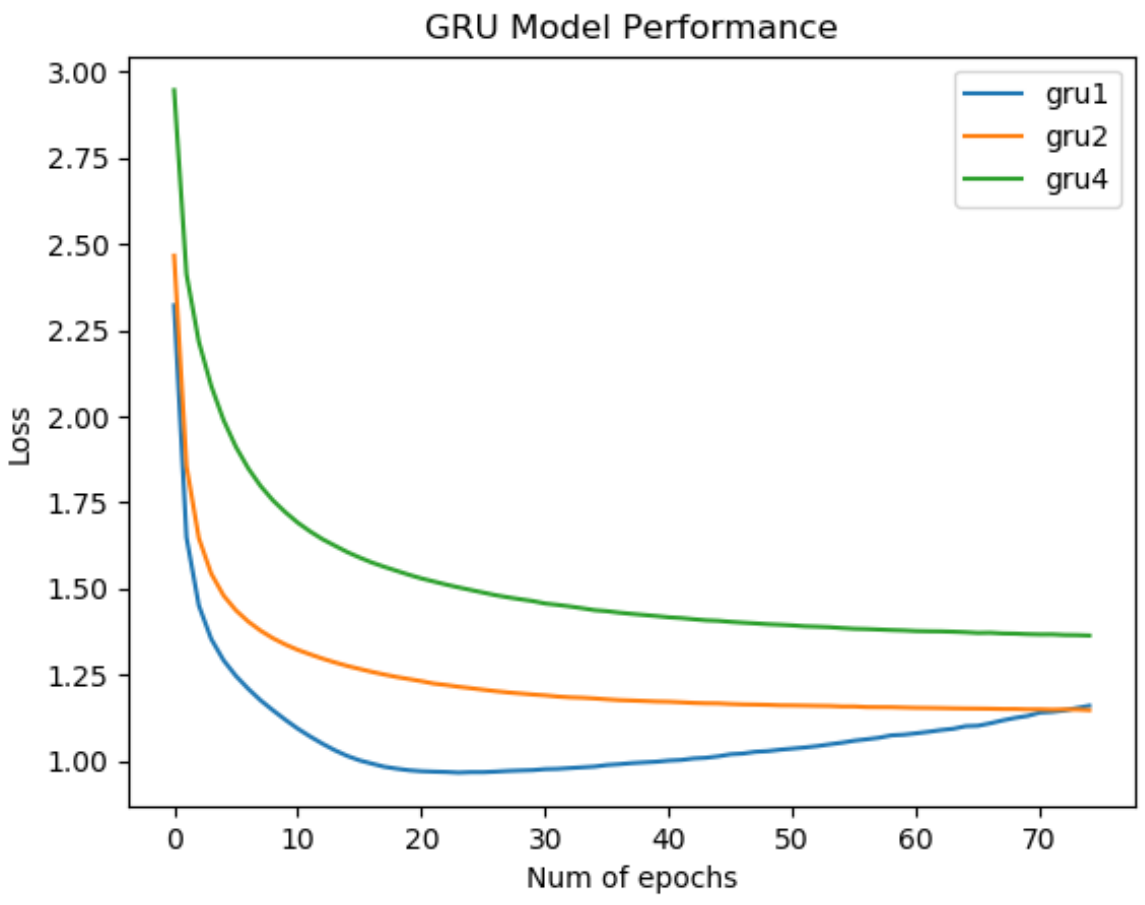

(b)

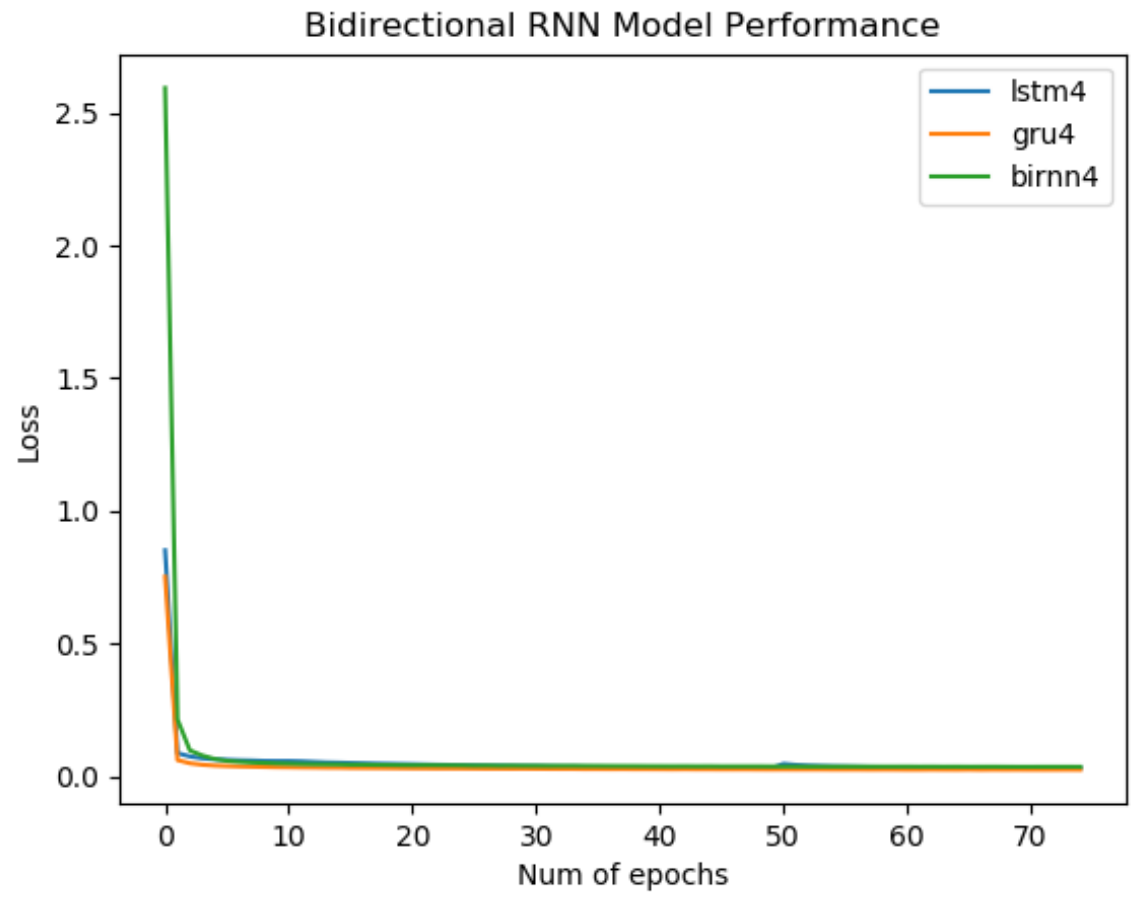

(c)

Fig. 5. An image describing all the losses for all intraneural layers, the image is a plot between the number of epochs (iterations) on the x-axis and losses on the y-axis. (a) represent a combination of all the LSTM layers and losses on each epoch. Blueline is for single-layered LSTM model, the orange line for the bi-layered model and green line for quad-layered Model. (b) is for describing GRU model performance with the blue line for the single-layered model, the orange line for the bi-layered model and green line for the quad-layered model. (c) describes model performance for BidirectionalRNN model with blue, orange and green lines for single-layered, bi-layered and quad-layered model.

On juxtaposing interneural layer performances for all the three algorithms in different architectures, many concealed ornamentations can be comprehended. Fig. 6 marks out these patterns. Fig. 6(a) is peculiar to single-layered model performance, embellishing anomalies shown by all the three algorithms when 1024 neurons are taken in a single layer. It is observed that LSTM and GRU show approximately the same performance around the 20th iteration. Later, the LSTM model tends to converge better and in Bidirectional RNN model, the losses deviate from these two models substantially. Bi-layered architecture and quad-layered architecture also exhibit the same pattern as of single-layered models with the only difference that LSTM and GRU architectures do not diverge much. Fig. 6(b) and fig. 6(c) show a vivid picture about these changes in bi-layer and quad-layer configuration.

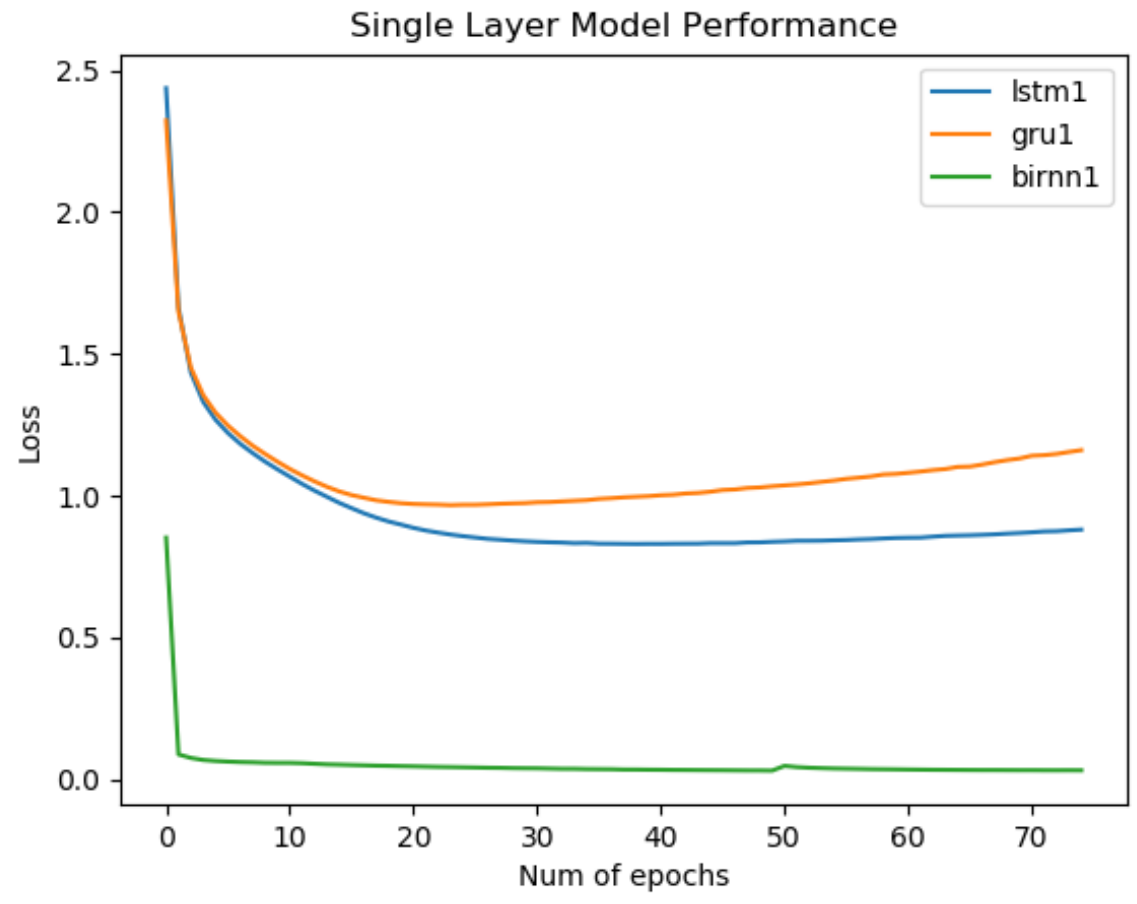

(a)

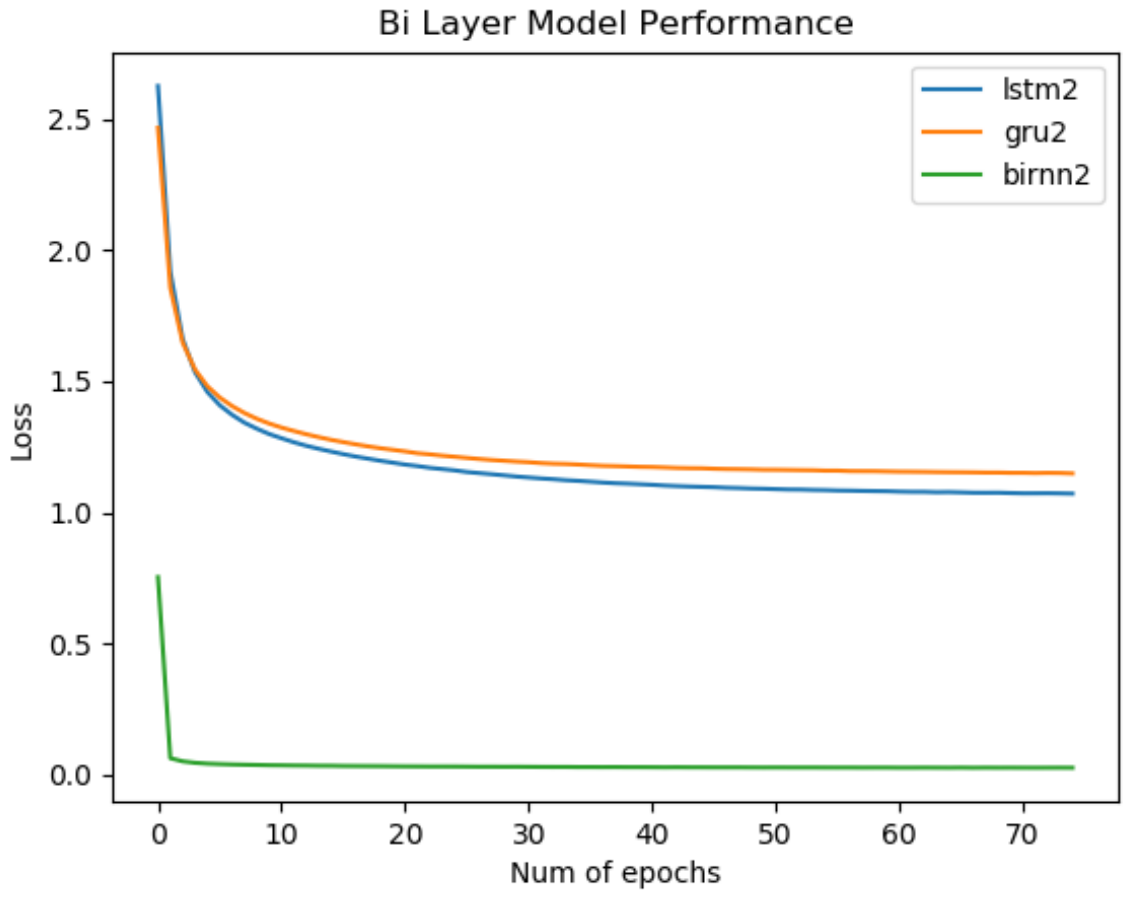

(b)

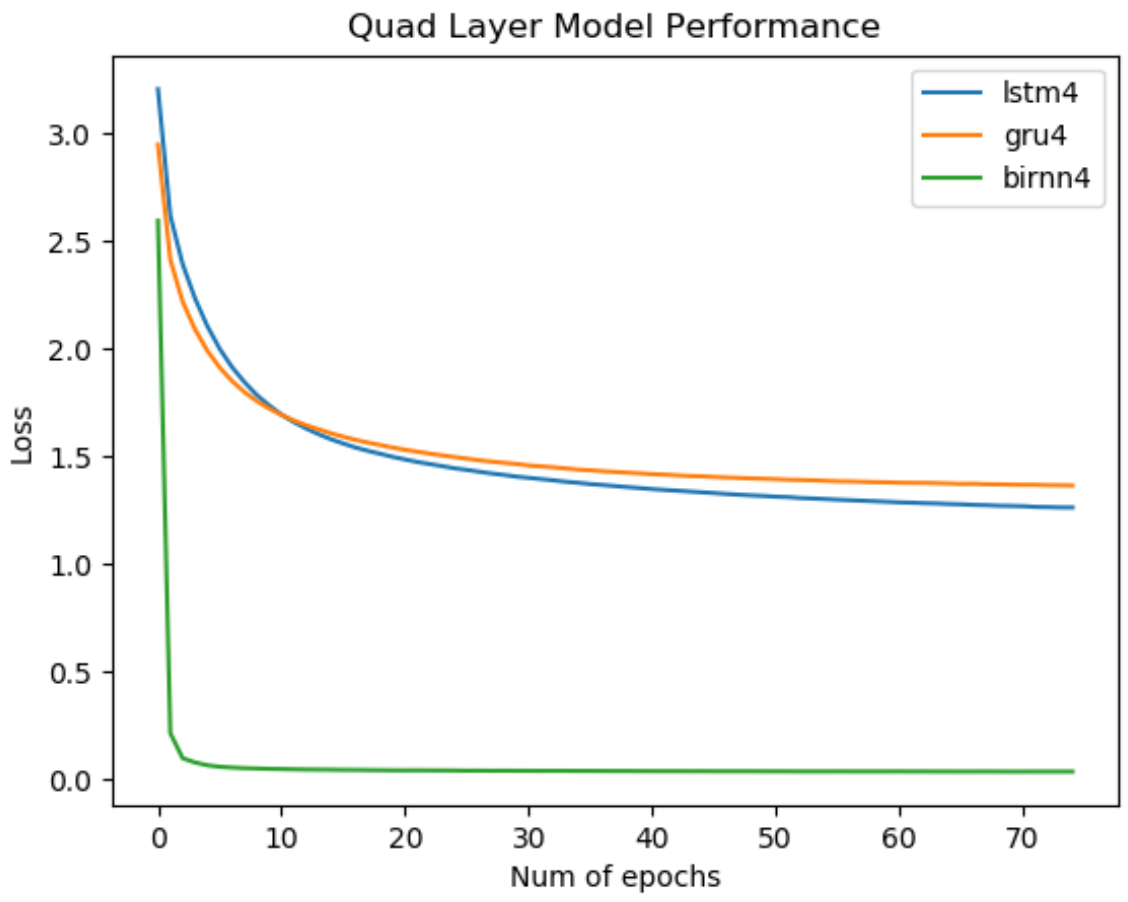

(c)

Fig. 6. The image describes interneural layer model performance for all the algorithms: LSTM, GRU and BidirectionalRNN. In all the images a plot is shown where X-axis represents the number of epochs (iterations) and losses on Y-axis. (a) describes single-layered model performance for all the three algorithms with blue, orange and green for LSTM, GRU and BidirectionalRNN respectively. (b) is for bi-layered models and (c) for quad-layered models. (b) and (c) both have the same plot properties as of (a)

Examples of some of the generated dialogues taking “JON:” as a starting input string after training all these models are shown in fig. 7. Though the results generated are understandable, they fail to learn some basic connectivity making it hard to comprehend the context and plot. For this case, only 1000 characters were generated for each architecture. More can be generated as per the need of the user. In fig. 7, only results from single-layered architecture are shown.

JON: I didn't work to discuss down the Prince Dorne, she nods.
MARWYN: The Lannisters have wed it in the skolloors.
CUT TO: OLDTOWN - CITADEL TOILDER: You're a good man.
YARA: I think we can win fighting for me. That's what I want.
We are leaving your name.
I did King's Landing now, I'm a bit worthy thing. For sure why not.
The Blackfish is the day I don't left our demmn sound at the Hall want to be here at this child, but it's a beggar, and you knew I imagine anymore. I beg you.
DAVOS walks away from YOUNG MAN #2 and JAIME.
BRONN: You don't gu's beat knowing what you should have vicaldered us.
DAVOS: Seize him. I benimated it was too bleeding. If you don't know anymore.
DAVOS: A shadow with a Tormund stairs to lock the horses. That's why the real of our king shits in a cure.
DAVOS. JON hands the arm.
DAENERYS: It’s your horde, amour wesdon't matter now.
JON: And that will you obey.
TORMUND: You're a bit orders.
JON: I'm sorry. You forgot the one who try north.

(a)

JON: Good. They will stand what was sweet desor.
I'm sorry.
ARYA would also still have me fear the word flast?
BRAN? Hodor hapes the moment then is in a larrer brothel.
OLENNA: With a wrong little chareea.
TYRION smiles at DRIGON.
ARYA: Her northerd talkers.
What do you want.
Tader to House will pay for our family. Now you want to meet you need you entay between here life.
They stop the perfowms raised, and she's unseen him.
DAVOS: We need to leave and set’s a wister is all aclorts crying.
ARYA turns back to BLAIMELY and his death and did be dead after confused.
YARA: I Have you matters. Mery don't like my father and chilking by the Commonernand.
TYRION: I demand to kill my soughing for.
She reason before they's command. Is it these to fight the court. I've ever believe that.
But that we'll go unoun.
ARYA: The Gods want to fight them off of your wines in he e1, the rest of the least.
CUT TO: VIRA’s room and bowhould come surviy the pams move for it, slaughteried for him.

(b)

JON:
JOJOJOJOJOJOJOJOJOJOJOJOJOJOJOJOJOJOJOJOJOJOJOJOJOJO
JOJOJOJOJOJOJOJOJOJOJOJOJOJOJOJOJOJOJOJOJOJOJOJOJOJO
JOJOJOJOJOJOJOJOJOJOJOJOJOJOJOJOJOJOJOJOJOJOJOJOJOJO
JOJOJOJOJOJOJOJOJOJOJOJOJOJOJOJOJOJOJOJOJOJOJOJOJOJO
JOJOJOJOJOJOJOJOJOJOJOJOJOJOJOJOJOJOJOJOJOJOJOJOJOJO
JOJOJOJOJOJOJOJOJOJOJOJOJOJOJOJOJOJOJOJOJOJOJOJOJOJO
JOJOJOJOJOJOJOJOJOJOJOJOJOJOJOJOJOJOJOJOJOJOJOJOJOJO
JOJOJOJOJOJOJOJOJOJOJOJOJOJOJOJOJOJOJOJOJOJOJOJOJOJO
JOJOJOJOJOJOJOJOJOJOJOJOJOJOJOJOJOJOJOJOJOJOJOJOJOJO
JOJOJOJOJOJOJOJOJOJOJOJOJOJOJOJOJOJOJOJOJOJOJOJOJOJO
JOJOJOJOJOJOJOJOJOJOJOJOJOJOJOJOJOJOJOJOJOJOJOJOJOJO
JOJOJOJOJOJOJOJOJOJOJOJOJOJOJOJOJOJOJOJOJOJOJOJOJOJO
JOJOJOJOJOJOJOJOJOJOJOJOJOJOJOJOJOJOJOJOJOJOJOJOJOJO
JOJOJOJOJOJOJOJOJOJOJOJOJOJOJOJOJOJOJOJOJOJOJOJOJOJO
JOJOJOJOJOJOJOJOJOJOJOJOJOJOJOJOJOJOJOJOJOJOJOJOJOJO
JOJOJOJOJOJOJOJOJOJOJOJOJOJOJOJOJOJOJOJOJOJOJOJOJOJO
JOJOJOJOJOJOJOJOJOJOJOJOJOJOJOJOJOJOJOJOJOJOJOJOJOJO
JOJOJOJOJOJOJOJOJOJOJOJOJOJOJOJOJOJOJOJOJOJOJOJOJOJO
JOJOJOJOJOJOJOJOJOJOJOJOJOJOJOJOJOJOJOJOJOJOJOJOJOJO
JOJOJOJOJ

(c)

Fig. 7. The image describes text generated by the trained model after taking “JON: “as an input. (a) represents text generated by single-layered LSTM model. (b) shows text generated by single-layered GRU model and (c) is for single-layered BidirectionalRNN model.

A complete list of files and text generated by this model can be found at [25].

## V. CONCLUSION

This paper achieves the goal of generating scripts from three different deep learning models. The models for text generation are trained using Bidirectional RNN, LSTM and GRU. The script of a famous TV series which contains all the dialogues’ and scenes’ description of all the episodes is successfully crunched into a single pickle file with the help of which the models are trained to generate the script of the next episode without any human intervention. The performance of the models is further analyzed to reach a conclusion that LSTM generates text a in most efficient way followed by GRU and then Bidirectional RNN while loss is least in Bidirectional RNN followed by LSTM and it is most in GRU. The LSTM model takes the least time for text generation, GRU takes slightly more time and Bidirectional RNN takes the highest time. The results obtained from different models are juxtaposed through the generated graphs and also, the scripts generated are presented effectively in this paper. Additionally, software implementation and related works in this domain are covered in different sections. We believe that further research can enhance this model and optimize it with lots of computation and data.

## ACKNOWLEDGEMENT

We would like to thank 4 “anonymous” reviewers for their so-called insights and their kind comments which helped in shaping early version of the manuscript, although any errors are our own and should not tarnish the reputation of these esteemed people.